\documentclass{article}
\usepackage{ijcai21}

\usepackage{times}

\usepackage{soul}
\usepackage{url}
\usepackage[hidelinks]{hyperref}
\usepackage[utf8]{inputenc}
\usepackage[small]{caption}
\usepackage{graphicx}
\usepackage{amsmath}
\usepackage{booktabs}
\usepackage{listings}
\usepackage{xcolor}

\definecolor{codegreen}{rgb}{0,0.6,0}
\definecolor{codegray}{rgb}{0.5,0.5,0.5}
\definecolor{codepurple}{rgb}{0.58,0,0.82}
\definecolor{backcolour}{rgb}{0.95,0.95,0.92}

\lstdefinestyle{mystyle}{
    backgroundcolor=\color{backcolour},   
    commentstyle=\color{codegreen},
    keywordstyle=\color{magenta},
    numberstyle=\tiny\color{codegray},
    stringstyle=\color{codepurple},
    basicstyle=\ttfamily\footnotesize,
    breakatwhitespace=false,         
    breaklines=true,                 
    captionpos=b,                    
    keepspaces=true,                 
    numbers=left,                    
    numbersep=5pt,                  
    showspaces=false,                
    showstringspaces=false,
    showtabs=false,                  
    tabsize=2
}

\title{AiTLAS: Artificial Intelligence Toolbox for Earth Observation}

\author{
Ivica Dimitrovski$^{1,2,*}$\and
Ivan Kitanovski$^{1,2}$\and
Pan\v{c}e Panov$^{1,3}$\and\\
Nikola Simidjievski$^{1,4}$\And
Dragi Kocev$^{1,3}$\footnote{Contact Authors}\\
\affiliations
$^1$Bias Variance Labs, Ljubljana, Slovenia\\
$^2$University of Ss Cyril and Methodius, FCSE, Skopje, N. Macedonia\\
$^3$Jo\v{z}ef Stefan Institute, Ljubljana, Slovenia\\
$^4$University of Cambridge, United Kingdom\\
\emails
\{ivica, ivan, pance, nikola, dragi\}@bvlabs.ai
}

\begin{document}

\maketitle

\begin{abstract}
The AiTLAS toolbox (Artificial Intelligence Toolbox for Earth Observation) includes state-of-the-art machine learning methods for exploratory and predictive analysis of satellite imagery as well as repository of AI-ready Earth Observation (EO) datasets. It can be easily applied for a variety of Earth Observation tasks, such as land use and cover classification, crop type prediction, localization of specific objects (semantic segmentation), etc. The main goal of AiTLAS is to facilitate better usability and adoption of novel AI methods (and models) by EO experts, while offering easy access and standardized format of EO datasets to AI experts which further allows benchmarking of various existing and novel AI methods tailored for EO data.



\end{abstract}

\section{Introduction}

According to the Sentinel Data Access Annual Reports~\cite{seninelreports}, the volume of data produced by the Sentinel satellite mission(s) is substantially growing – doubling each year since 2017 (to 19.5 PB in 2019), with data downloads growing at a similar rate (to 178.5 PB in 2019). 
On the other hand, this growth rate in data production is well matched by the rapidly growing development in AI, which probes various aspects of natural sciences, technology and society. 
Recent trends in machine learning, and particular deep learning, have ushered to a new era of image analysis and raised the predictive performance bar in many application domains, including remote sensing and Earth observation \cite{Ball17:jrnl}. 
Nevertheless, despite the success of some individual attempts, given the amount of available satellite images, the application of AI in the EO supply chain is still scarce and, more importantly, still unstructured and unverifiable. 
This is due to a number of challenges that require immediate attention: (1) lack of benchmark data sets produced in an AI-ready format, (2) lack of data and methods for benchmarking and model comparisons, and (3) lack of ready-to-use, comprehensive resources.

AiTLAS\footnote{{\color{blue}\url{https://github.com/biasvariancelabs/aitlas}}} addresses the aforementioned challenges and aims at resolving them. It is a recent open-source library designed to facilitate the use of EO data in the AI community and, more importantly, to accelerate the uptake of (advanced) machine learning methods and approaches by EO experts. More specifically, AiTLAS provides resources such as: benchmarking tools, ready-to-use models, tools for learning models de novo, semantically annotated datasets created in a format that can be used directly by AI methods. 

\section{AiTLAS}

We designed and developed a toolbox that contains state-of-the-art AI methods and techniques and follows the best practices for machine learning use and deployment. 
The toolbox is based on three main concepts: \textit{models}, \textit{datasets} and \textit{tasks}. The \textit{models} concept defines the architecture of the deep learning models as well as their behavior. The \textit{datasets} concept defines the data by encapsulating certain operations over datasets, such as loading and preparing the data as well as acquiring and, optionally, transforming a given item. Finally, the \textit{task} concept is used to make a complete workflow where a given model is used over a given dataset. In the toolbox, we have defined several tasks that are common for these types of problems, such as, tasks for \textit{training} and \textit{evaluating} a model over a given dataset, etc. 

Additionally, AiTLAS also includes \textit{transformations}, \textit{visualizations} and \textit{evaluation measures}, as supportive concepts to the previous ones. In particular, various \textit{transformations} can be applied to the data if needed before training. The \textit{evaluation measures} are used to gauge the performance of the model (e.g., accuracy, F1 score etc.) that, depending on the tasks, can also be illustrated with a variety of \textit{visualizations} (e.g., showing mask overlays makes sense for segmentation). Note that, for increasing usability, the evaluation measures, along with the loss functions, are logged in a Tensorboard~\cite{tensorflow2015-whitepaper} friendly format. 

\subsection{Software implementation}

AiTLAS is developed using \texttt{Python} and distributed under the MIT license. It is developed as a library but can be used as a standalone application as well. All the models, datasets and tasks can be configured via a proprietary JSON format or programatically by initializing the appropriate classes. The toolbox has a variety of dependencies that are used to power it, but the main one is PyTorch \cite{NEURIPS2019_9015}. This serves as the backbone over which deep learning models and behavior are built. 

All the functionalities of AiTLAS are organized into five separate modules: 
 \begin{itemize}
    \item \texttt{aitlas.base} - The core module that contains the abstract definitions of everything we have in AiTLAS. The definitions of the tasks, models, datasests, and transformations, evaluation measures and visualizations. To extend the functionality of AiTLAS, one would need to extend the definitions implemented here. 
    \item \texttt{aitlas.models} - Contains specific implementations of the deep learning models.
    \item \texttt{aitlas.datasets} - Contains The datasets and dataset types which are supported by AiTLAS, out of the box. 
    \item \texttt{aitlas.tasks} - Includes readily executable workflows. There are several implemented tasks in AiTLAS, such as: training task, evaluating tasks, task to run some data preparations, tasks for extracting features etc. While the ones currently implemented are can be applied in many different scenarios, they can also serve as a blueprint for creating/instantiating new, more specific, task.
    \item \texttt{aitlas.utils} - Contains handy utility functions, which can be used within the toolbox or outside of it. 
\end{itemize}

\subsection{Methods and models}

The toolbox supports a number of models that have been shown to lead to improved predictive performance. Its flexible design of the software architecture allows users to easily include other models, including implementing their own custom made models. AiTLAS currently includes the following models: DeepLabv3 \cite{chen2017rethinking}, Fast R-CNN \cite{girshick2015fast}, ResNet \cite{he2015deep}, VGG16 \cite{simonyan2015deep}, and Unsupervised DeepCluster \cite{caron2019deep}. 

The implemented models cover a variety of EO use cases: land use and land cover classification, semantic segmentation and object detection in the context of EO, etc. The models support both multi class and multi label classification. The multi class models assign a single label per image, whereas the multi label models assign multiple labels (or, in practice, land covers) in each image, which is closer to what is used in real life. 

The Unsupervised DeepCluster \cite{caron2019deep} model can leverage large amounts of unlabeled images to learn robust features. It uses clustering in combination with deep neural networks to provide pseudo-labels for a convolutional neural network. Within the toolbox, we are using the unsupervised deep learning to learn general features from a large set of unlabeled satellite images and then apply fine-tuning using datasets with a limited number of labeled images for a given task.

\subsection{Datasets}

AiTLAS has a number of datasets available. We have separated them into two groups: (i) land-use/cover classification datasets and (ii) object segmentation datasets. Note that, land use datasets are where the annotation problem is a multi class or a multi label type, where as the latter tackles segmentation. To this end, the former includes:  UC Merced \cite{yang2010bag,chaudhuri2017multilabel}, Eurosat~\cite{helber2019eurosat}, BigEarthNet~\cite{sumbul2019bigearthnet},  DFC15 \cite{hua2019recurrently},  PatternNet \cite{zhou2018patternnet}, AID \cite{xia2017aid} and RESISC45 \cite{Cheng_2017}. The latter includes: Chactun \cite{Somrak20:jrnl}, SpaceNet 6 \cite{dukai_2018} and Landcover AI \cite{boguszewski2020landcoverai}. 





\section{AiTLAS use case}

In order to demonstrate the functionalities of the toolbox, we present a Jupyter Notebook for multi label land cover classification of satellite images. We will show how to load the data and inspect the classes and class imbalances in the dataset. Next, we train our model using convolution neural networks, and finally, we test the model with external images for inference.

We use the multi label UC Merced dataset with 17 land cover classes. Once we get the data and unzip it\footnote{\url{https://bigearth.eu/datasets.html}}, we are ready to explore it. In order to achieve this we need to create and instance of the UcMercedMultiLabelDataset, which is a class within the toolbox that can handle loading of the data. The instance is created by providing the root folder in which the images and annotations are unzipped, additionally we can supply the batch size to be used in the data loader during the process of training, the shuffle parameter to have the data reshuffled at every epoch and number of workers to specify the subprocesses to use for data loading. As an illustration the code snippet is given in Listing~\ref{lst:load_data}. 

\begin{lstlisting}[language=Python, label={lst:load_data}, caption=Load train dataset]
train_dataset_config = {
    "batch_size": 16,
    "shuffle": True,
    "num_workers": 4,
    "root": "/ucmerced/images"
}
train_dataset = UcMercedMultiLabelDataset(train_dataset_config)
fig = train_dataset.show_image(340)
\end{lstlisting}

\begin{figure}[!b]
\includegraphics[width=0.35\textwidth]{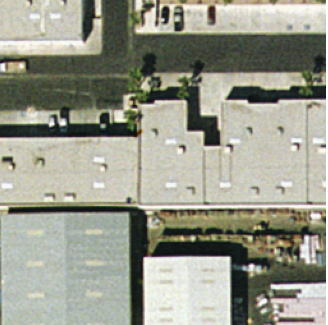}
\centering
\caption{Example image from the UC Merced multi-label dataset (Image\#340) for which the predictions are 'buildings', 'cars', 'pavement' and 'trees'. 
}
\label{fig:sample_image}
\end{figure}

To display a given image from the dataset you can use the function {\tt show\_image} implemented in the class for the dataset. The function expects an index/order number and will display the image and the annotations/labels for the image. The example image with index/order number 340 and the annotations/labels are shown on Figure~\ref{fig:sample_image}. 

Before we move on to the machine learning task, we can inspect the distribution of the classes in the dataset. Checking the distribution of the dataset is an important step to check for data imbalances in your dataset. The distribution can be easily calculted using the function {\tt data\_distribution\_table} implemented in the class for the dataset. The distribution for the dataset is given in Figure~\ref{fig:sample_image}. The data reveals the class imbalances in the dataset, the pavement class has $987$ images while Airplane class has $78$ images.

Next, we need to initialize and create the model with configuration. We will use the ResNet50 architecture which is implemented in the toolbox. The configuration parameters are the number of epoch, the path in which to save the final model parameters, the visualizations and the results, the number of classes, the learning rate, the threshold to obtain the predictions and calculate the various metrics. By setting the parameter pretrained to $true$ we state that we want to use the pretrained variant of ResNet50 over the ImageNet dataset and to apply fine tuning using the UC merced multi label dataset. Now, we can start training our model with the data. We use the {\tt train\_and\_evaluate\_model} function. The function accepts the train dataset created using the code from Listing~\ref{lst:load_data}, similar to this we can create the test dataset using different root folder (UC Merced multi label dataset has predefined train and test splits). We can also select to apply transformations over the images, this can be configured using the parameters {\tt transforms} and {\tt target\_transforms} defined for the dataset class. As an ilustration the code snippet for initializing the model and initiating the training is given in Listing~\ref{lst:train_model}. 

\begin{lstlisting}[language=Python, label={lst:train_model}, caption=Creating a model and start of model training]
epochs = 50
model_directory = "/ucmerced/experiments/"
model_config = {"num_classes": 17, "learning_rate": 0.0001,"pretrained": True, "threshold": 0.5}
model = ResNet50MultiLabel(model_config)
model.prepare()
model.train_and_evaluate_model(
    train_dataset=train_dataset,
    epochs=epochs,
    model_directory=model_directory,
    val_dataset=test_dataset,
    run_id='1',)
\end{lstlisting}

Once the training starts, the AiTLAS toolbox displays the running time for each epoch, calculated loss and the evaluation metrics. The final model will be saved in the model directory given in the configuration parameter. The calculated metrics and losses are logged and can be visualized using \textit{tensorboardX}.
To test the model, we can predict several images from an external source and see how the model performs. This can be done using the code from Listing~{\ref{lst:predict_model}}. First we load the image using the utility function {\tt image\_loader} from the toolbox, load the model parameters from the saved file and call the predict function. The predict function returns the predicted labels and the probability for each label. 

\begin{lstlisting}[language=Python, label={lst:predict_model}, caption=Loading trained model and predicting image]
model.load_model('ucmerced/experiments')
image = image_loader('images/predict')
plt.imshow(image)
y_true, y_pred, y_prob = model.predict_image(image)
\end{lstlisting}

The entire code and Jupyter Notebook for this tutorial is available in the Github repository of the AiTLAS toolbox\footnote{\url{https://github.com/biasvariancelabs/aitlas/tree/master/examples}}.

\section{Summary}

We have described AiTLAS, an open-source, state-of-the-art toolbox for exploratory and predictive analysis of satellite imaginary pertaining to a variety of different tasks in Earth Observation. AiTLAS has several distinguishing properties. First, it is modular and flexible - allowing for easy configuration, implementation and extension of new data and models. Next, it is general and applicable to a variety of tasks and workflows. Finally, it is user-friendly. This, besides aiding the AI community by providing access to structured EO data, more importantly, facilitates and accelerates the uptake of (advanced) machine learning methods by the EO experts, thus bringing these two communities closer together.

\section*{Acknowledgments}
The development of the AiTLAS toolbox is supported by the grant from the European Space Agency (ESRIN): AiTLAS - Artificial Intelligence toolbox for Earth Observation (ESA RFP/3-16371/19/I-NB).

\clearpage

\bibliographystyle{named}
\bibliography{aitlas_demo}

\end{document}